# Time-EAPCR: A Deep Learning-Based Novel Approach for Anomaly Detection Applied to the Environmental Field

Lei Liu[1,2], Yuchao Lu[1,2], Ling An[1,2], Huajie Liang[1,2], Chichun Zhou[1,2,] *, and Zhenyu Zhang[1,2,]*

[1]School of Engineering, Dali University, Yunnan, 671003, China

[2]Air-Space-Ground Integrated Intelligence and Big Data Application Engineering Research Center of Yunnan Provincial Department of Education, Yunnan, 671003, China

*Corresponding author: Zhenyu Zhang: zhangzhenyu@dali.edu.cn

Chichun Zhou: zhouchichun@dali.edu.cn

#These authors contributed equally to this work

Emails of other authors: Lei Liu: liulei@stu.dali.edu.cn

Yuchao Lu : luyuchao@stu.dali.edu.cn

Ling An: anling@stu.dali.edu.cn

Huajie Liang :lianghuajie@stu.dali.edu.cn

# Abstract

As human activities intensify, environmental systems such as aquatic ecosystems and water treatment systems face increasingly complex pressures, impacting ecological balance, public health, and sustainable development, making intelligent anomaly monitoring essential. However, traditional monitoring methods suffer from delayed responses, insufficient data processing capabilities, and weak generalisation, making them unsuitable for complex environmental monitoring needs.In recent years, machine learning has been widely applied to anomaly detection, but the multi-dimensional features and spatiotemporal dynamics of environmental ecological data, especially the long-term dependencies and strong variability in the time dimension, limit the effectiveness of traditional methods.Deep learning, with its ability to automatically learn features, captures complex nonlinear relationships, improving detection performance. However, its application in environmental monitoring is still in its early stages and requires further exploration.This paper introduces a new deep learning method, Time-EAPCR (Time-Embedding-Attention-Permutated CNN-Residual), and applies it to environmental science. The method uncovers feature correlations, captures temporal evolution patterns, and enables precise anomaly detection in environmental systems.We validated Time-EAPCR's high accuracy and robustness across four publicly available environmental datasets. Experimental results show that the method efficiently handles multi-source data, improves detection accuracy, and excels across various scenarios with strong adaptability and generalisation. Additionally, a real-world river monitoring dataset confirmed the feasibility of its deployment, providing reliable technical support for environmental monitoring.

# 1 Introduction

With the rapid development of industrialisation and urbanisation, human activities have increasingly impacted the natural environment, placing unprecedented pressure on environmental systems. The combination of multiple pollution sources, such as industrial wastewater discharge, agricultural non-point source pollution, and domestic sewage, has made the environmental pollution issue more complex and severe (Conrad & Hilchey, 2011; Yan et al., 2011). Additionally, urban water systems are also under immense pressure (Bacco et al., 2017). These systems not only face traditional challenges such as water quality pollution and facility ageing, but also need to cope with extreme weather events and flood risks exacerbated by climate change. These issues directly affect public health, safety, and sustainable social development (Barbosa et al., 2012; Johnson & Munshi-South, 2017). Therefore, intelligent environmental anomaly detection technologies have become a critical tool for ensuring ecological security and promoting sustainable development (Zaghloul et al., 2020).

Anomaly detection plays a crucial role in environmental monitoring. In recent years, with the rapid development of the Internet of Things (IoT) (Lazarescu, 2013), edge computing (Roostaei et al., 2023), high-precision sensors (Pule et al., 2017), and automated monitoring equipment (Dunbabin & Marques, 2012), environmental monitoring systems have gradually gained the capability for real-time perception and data transmission. As a result, efficient anomaly detection methods have become increasingly important.Earlier environmental anomaly detection methods primarily relied on mathematical modelling techniques such as Bayesian networks (Hill et al., 2007), ESAD (Bezdek et al., 2011), and PE (Liu et al., 2015), which were widely used for natural disaster detection, marine environment monitoring, and water treatment system anomaly identification. However, these methods often rely on linear assumptions or specific distributions (e.g., Gaussian distribution) (Erickson et al., 2015), making them less effective in addressing the complex nonlinear relationships and high-dimensional spatiotemporal characteristics present in environmental data. This limitation in generalisation capabilities has prompted researchers to explore more intelligent techniques.

In recent years, with the rapid increase in the scale of environmental data and computational capabilities, data-driven artificial intelligence (AI) technologies have shown great potential in the field of environmental monitoring (Konya & Nematzadeh, 2024). An increasing number of studies have explored the intelligent application of machine learning techniques in environmental monitoring. For example, Georgescu et al. (Georgescu et al., 2023) proposed a hybrid model based on cascade forward networks and random forests, which improved water quality prediction accuracy.

Hammond et al. (Hammond et al., 2021) evaluated the performance of SVM, random forests, MLP, and gradient boosting models for sewage treatment anomaly classification. Abokifa et al. (Abokifa et al., 2019) combined principal component analysis (PCA) and artificial neural networks (ANNs) for network attack anomaly detection in urban water supply systems. Chen et al. (Chen et al., 2020) conducted a systematic analysis of CNN models in the field of water quality prediction. These studies highlight the broad prospects for the application of artificial intelligence in environmental monitoring.

Environmental monitoring data typically includes time-series features such as trends, seasonal patterns, and change points, which reveal the dynamic evolution of environmental variables and are crucial for anomaly detection in system states. In recent years, a series of research achievements have been made in the field of intelligent monitoring technologies for time-series data. For instance, Choi et al. (Choi et al., 2021) systematically summarised time-series anomaly detection methods based on artificial intelligence, highlighting the key role of time-series data in system monitoring. Renaud et al. (Renaud et al., 2023) evaluated the application of LSTM and Transformer models in the field of noise monitoring, confirming the advantages of these models in identifying transient pulse interference and gradual pollution. Ansari et al. (Ansari et al., 2018) proposed an ARIMA-NAR hybrid model for forecasting the inflow and outflow of wastewater treatment systems, providing an effective technical pathway to enhance wastewater treatment efficiency.

The multi-source sensor data collected by environmental monitoring systems possess complex characteristics, such as high-dimensional coupling, nonlinear dynamic interactions, and temporal dependencies, which pose significant challenges for feature extraction and anomaly detection. Unlike image and text data, which typically exhibit explicit feature correlation patterns and prior information, environmental data often lacks directly observable, explicit patterns of variable relationships. For example, the relationship between meteorological parameters and pollutant dispersion is often influenced by complex environmental factors, making it impossible to describe with simple linear relationships (Karpatne et al., 2017).Moreover, multi-source monitoring data not only contain linear correlations, but more commonly exhibit higher-order nonlinear interactions, such as the dynamic coupling relationship between pH and chemical oxygen demand (COD) in water quality monitoring, which demonstrates non-monotonic and threshold-sensitive complex responses (Liu et al., 2021).

On the other hand, the complex dynamical characteristics of environmental systems are not only reflected in individual time series but also involve cross-dimensional nonlinear temporal coupling across multi-source heterogeneous data. For instance, the temporal variations in dissolved oxygen (DO) and nitrogen-phosphorus concentrations can provide critical insights into the state of aquatic ecosystems, revealing their nonlinear dynamic interactions (Su et al., 2022).

However, traditional machine learning methods, such as linear regression, support vector machines (SVMs), and random forests, exhibit significant limitations when processing high-dimensional, nonlinear temporal data. These methods typically rely on linear assumptions or local approximations, making it difficult to capture implicit dynamic relationships within the data and resulting in weak generalisation capabilities for complex environmental systems.At present, deep learning methods, such as Convolutional Neural Networks (CNNs) and Long Short-Term Memory (LSTM) networks, still face challenges in feature extraction when applied to environmental data. They struggle to effectively capture the high-dimensional coupling, nonlinear dynamic interactions, and temporal dependencies inherent in such datasets. Therefore, there is an urgent need for more advanced deep learning techniques that can seamlessly integrate temporal modelling with feature extraction capabilities, enabling them to better address the challenges posed by the high dimensionality, nonlinearity, and temporal dependencies of environmental data.

To address the aforementioned technical challenges and meet the practical engineering demands of environmental monitoring, this study proposes an innovative deep learning-based framework, Time-EAPCR (Time-Embedding-Attention-Permutated CNN-Residual). This model achieves optimised anomaly detection for multi-source environmental sensor data by deeply integrating spatiotemporal feature encoding with cross-dimensional relational inference.The key technological advancements of this approach are reflected in two core design principles:1)Multi-source environmental parameter association modelling – By leveraging a deep learning module to extract features from multi-dimensional sensor data at the same time step, this approach effectively captures cross-sensor dependencies and correlations within multi-source environmental data.2)Time series feature association modelling – Through a feature interaction mechanism, the model establishes potential relationships across different features in the temporal dimension, thereby identifying critical time series interaction patterns essential for environmental monitoring.

To verify the effectiveness of the proposed method, benchmark tests were conducted on four publicly available environmental system datasets, supplemented by transfer validation using real-world environmental monitoring data. Cross-scenario experiments demonstrate that the proposed approach exhibits superior performance in both natural water environments (e.g., rivers and ponds) and urban infrastructure systems (e.g., underground pipelines and wastewater treatment facilities), while maintaining robust anomaly detection capabilities under complex environmental

dynamics.Based on our work, the following two key contributions can be summarised:

1) A deep learning-based Time-EAPCR method is proposed, which integrates time series modelling with multi-dimensional feature interaction analysis. This approach effectively captures both temporal dependencies and complex inter-feature relationships within environmental data, significantly enhancing anomaly detection accuracy and robustness.

2) A highly generalisable and adaptable anomaly detection framework is constructed. Experimental validation on publicly available datasets across diverse scenarios, including natural ecosystems and urban environmental systems, demonstrates the universality and efficiency of the proposed method in complex environmental settings. This research provides a novel technical pathway for advancing intelligent solutions in environmental monitoring.

## 2 Materials and methods

### 2.1 Dataset description

### 2.1.1 Public dataset

To assess the effectiveness of Time-EAPCR in environmental anomaly monitoring, we conducted validation using datasets that are publicly available in relevant literature on environmental anomaly detection. These datasets include three urban water environment systems and one natural environment system (Russo et al., 2021).The selected datasets comprehensively represent the typical challenges encountered in environmental monitoring, such as sensor failures in urban water circulation systems and random disturbance events in natural environments, which introduce real-world interference factors. Additionally, these datasets exhibit multi-dimensional feature variables, varying temporal resolutions, and spatiotemporal dependencies, making them well-suited for evaluating the robustness and generalisability of the proposed method. A detailed statistical comparison of the datasets is presented in Table 1.

The Eawag Ponds dataset originates from long-term ecological monitoring of pond ecosystems in Dübendorf, Switzerland. It comprises 16 controlled experimental ponds (hereafter referred to as ponds) designed to simulate a multi-variable interactive ecosystem by introducing different species of aquatic plants and molluscs as biological regulatory factors (Narwani et al., 2019).This dataset features high-frequency time-series data collected at 15-minute intervals over a period of 234 days. Each experimental pond is equipped with eight environmental parameter sensors, capturing key water quality indicators, including electrical conductivity, chlorophyll fluorescence, phycocyanin fluorescence, dissolved organic matter fluorescence, dissolved oxygen, pH, and temperature.To facilitate a supervised learning framework, the research team, in collaboration with domain experts, manually annotated anomalies within 7,488 hours (24,464 sampling points) of raw data. The final standard test set contains 2% anomalous samples, encompassing two primary types of anomaly events: equipment failures and ecological disruptions.

UWO S1 and UWO S2 datasets originate from a long-term sewer process monitoring programme conducted in Fehraltorf, Zurich, Sweden. These time-series datasets were collected over two periods: March 26 to April 25, 2017, and September 1 to November 12, 2017, respectively.Both datasets contain manually annotated anomalies identified by domain experts, with UWO S1 exhibiting relatively simple contextual anomaly patterns, whereas UWO S2 presents more complex collective anomaly patterns.

The WaterHub dataset originates from an urban wastewater sustainability treatment research project, systematically capturing the full-process data of wastewater treatment from building drainage units (e.g., washbasins and showers). The monitoring period spans 10 months (approximately 304 days).The treatment system employs a dual-stage process integrating a membrane bioreactor (MBR) and a biological activated carbon (BAC) filter. Two pressure sensors are deployed in the MBR unit to monitor biofilm dynamics, while two additional pressure sensing nodes in the BAC unit track adsorption efficiency, forming a four-dimensional sensor matrix.A domain expert-driven multi-dimensional analytical framework was utilised to identify five typical anomaly types:1) MBR foam accumulation 2) Membrane module clogging 3) MBR liquid level switch failure 4) BAC pressure drop anomaly 5) Data acquisition interruptions. The benchmark dataset, constructed through a rigorous annotation process, contains 436,320 monitoring records, with anomalies accounting for 13.3% of the data. It comprehensively represents multi-scale fault patterns in wastewater treatment systems, ranging from physical blockages to data-level anomalies.

**Table 1**. Statistical Overview of the Four Public Datasets

| Datasets | Observations | Variables | Anomalies (%) |
| --- | --- | --- | --- |
| Ponds | 22,464 | 8 | 2.34 |
| UWO S1 | 14545 | 3 | 16.2 |
| UWO S2 | 209,79 | 3 | 27.0 |
| WaterHub | 436320 | 4 | 13.3 |

It is important to highlight that anomalous events in environmental monitoring scenarios occur infrequently, leading to a general scarcity of anomaly samples across datasets. Within a supervised learning framework, the impact of data distribution on model performance primarily manifests as a class imbalance problem. As shown in

Table 1, except for UWO S2 dataset, the proportion of anomaly samples in the remaining datasets is below 20%.To address this issue, the following data preprocessing strategies were applied to the training and test sets:The original distribution of the test set was preserved to accurately reflect real-world scenarios.For training sets where anomaly proportions were below 20%, an adaptive sample replication-based oversampling strategy was implemented. Using temporal shift augmentation, the anomaly sample ratio was increased to 20% (i.e., a 4:1 normal-to-anomaly sample ratio). This approach mitigates class imbalance while effectively reducing the risk of overfitting.

### 2.1.2 Private dataset

To evaluate the applicability of the proposed method in real-world engineering environments, this study additionally constructed a water quality monitoring dataset for the Luoshi River Basin in Dali, Yunnan. As a key inflow channel to Erhai Lake (monitoring locations shown in Figure 1), the water quality dynamics of this basin directly impact the ecological security of Erhai Lake. The dataset covers a monitoring period from November 2022 to November 2024.A total of nine in-situ sensors were deployed to synchronously collect nine key physicochemical parameters, including total nitrogen (TN), total phosphorus (TP), ammonia nitrogen ($NH_3$-N), water temperature, pH, chemical oxygen demand (COD), electrical conductivity, dissolved oxygen (DO), and turbidity. Each sensor recorded measurements at 4-hour intervals throughout the monitoring period.The preprocessing of the raw data involved two key steps:1) Handling missing data: To address discontinuities in the collected data, a time-series linear interpolation method was applied. This approach estimates missing values through linear fitting between adjacent valid data points, ensuring a reasonable reconstruction while preserving the temporal evolution patterns of water quality parameters.2) Data annotation based on regulatory standards: Water quality classification was conducted in accordance with the Chinese National Surface Water Quality Standard (GB 3838-2002), which categorises water quality into five classes (Ⅰ–Ⅴ), ranging from high to low quality. Class V represents water suitable for agricultural irrigation and general landscape use, while water quality exceeding Class V is considered heavily polluted.According to Dali Municipal Government's assessment, the water quality of the Luoshi River has generally remained at or above Class IV over the past two years, though temporary occurrences of Class V water quality have been observed during the rainy season. Consequently, we defined sub-Class V water bodies (i.e., those exceeding Class V pollution thresholds) as anomalous events, which were used for data annotation.The final dataset comprises 4,130 records across nine feature dimensions, including 483 anomalous samples, accounting for 11.6% of the total data.

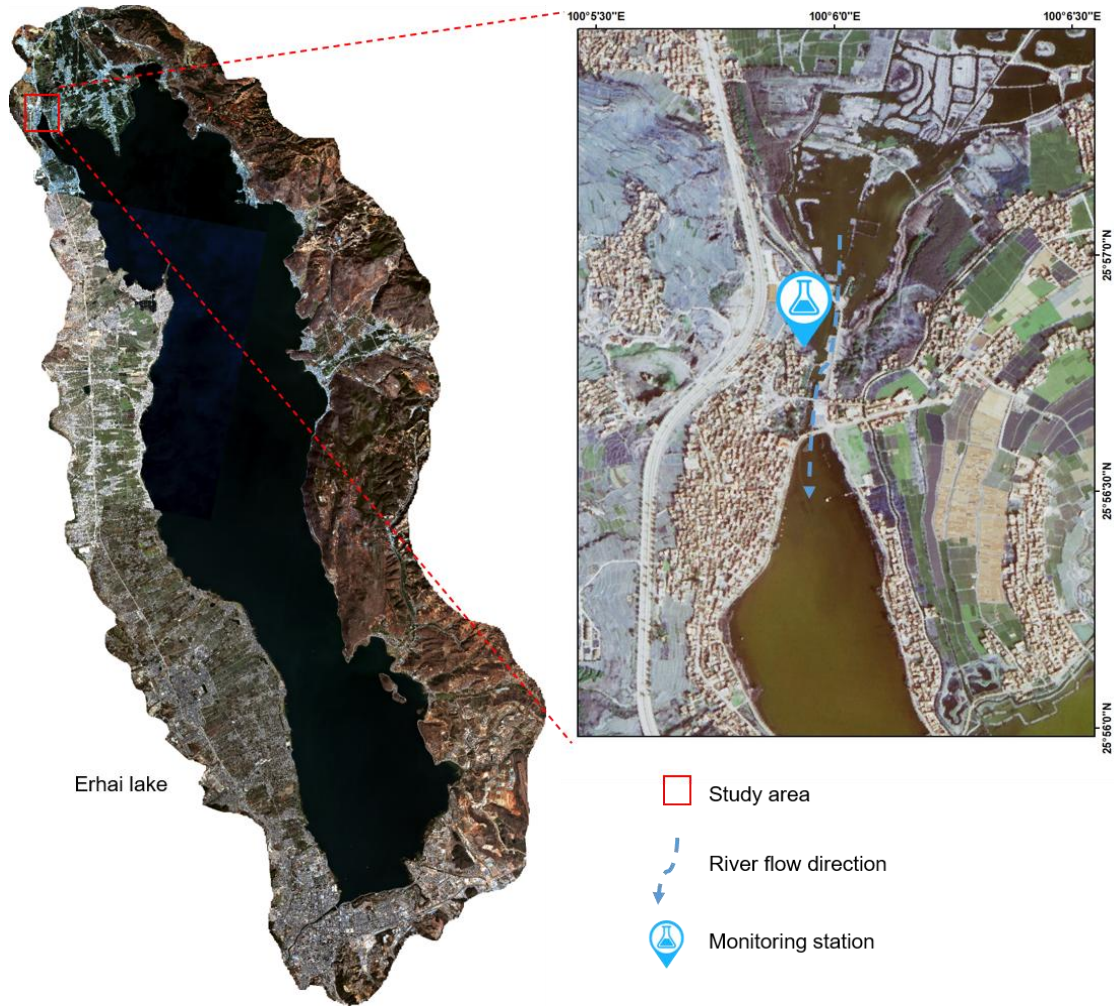

**Figure 1.** Schematic Diagram of Monitoring Station Locations for Private Data Sources

## 2.2 Method

### 2.2.1 Problem statement

Environmental monitoring systems continuously collect multi-dimensional time-series data through N sensors over T time points, where each observation at time t can be represented as a feature matrix. The correlation among multiple sensors is crucial for effective feature extraction.However, unlike traditional image, graph, or text data, multi-source environmental data typically lack explicit relational structures. As a result, conventional feature extraction modules (such as CNNs and GNNs) that rely on explicit topological relationships struggle to be effectively applied in multi-variate time-series scenarios where no prior structural constraints exist. Therefore, a key challenge in achieving precise anomaly detection lies in effectively capturing the implicit correlations among features.

Additionally, each feature exhibits a unique temporal dependency, which can be captured by applying a sliding window to extract its historical sequence over w time points, thereby modelling the univariate temporal evolution pattern.However, this approach solely focuses on the temporal variations within individual features and lacks the capability to model cross-dimensional temporal correlations between different features. Therefore, it is essential to introduce more effective methods for computing temporal dependencies across features, thereby enhancing the ability to characterise the dynamic evolution patterns in environmental data.

In summary, achieving accurate anomaly detection requires a comprehensive consideration of both intra-sample feature correlations and temporal dependencies across different samples, enabling a more holistic capture of anomalous patterns. There is an urgent need for novel deep learning architectures to overcome these challenges.

## 2.2.2 Method introduction

We propose Time-EAPCR to address the challenges associated with multivariate time-series anomaly detection in environmental systems. The method consists of two primary components: multi-sensor feature processing and time-series processing. These components are further divided into five functional modules:1) LSTMs-TFE (Time Feature Extraction) (Yu et al., 2019);2) Embedding Process (Mikolov et al., 2013);3) Bilinear Attention (Kim et al., 2018) for constructing the correlation matrix;4) Permute CNN;5) Residual Connection for sampling feature correlations.Among these, LSTMs-TFE is applied to the time-series processing component, while the embedding process is utilised in the raw point feature processing component. The remaining modules are involved in both processing components. These details will be elaborated in the following sections.

Multi-Sensor Feature Module:The core objective of this module is to establish association patterns among multi-source environmental parameters. Before inputting the data into the model, feature discretisation is applied, where raw sensor data are mapped to discrete categorical identifiers based on predefined threshold intervals. This process reduces computational complexity while preserving information integrity. Given that sensor data typically exhibit high granularity, carefully adjusting classification thresholds does not introduce significant negative effects.To standardise the data format, the Embedding process is applied, followed by the Bilinear Attention mechanism, which explicitly captures correlation patterns between features. Subsequently, CNN and Permute CNN perform extensive feature sampling, while Residual Connection integrates outputs from different feature extraction modules. This design enhances the model's generalisation capability and effectively extracts associative patterns across diverse sensor data.

Embedding Process: At each time step, discretised integer identifiers of different features along the temporal dimension are mapped into high-dimensional feature vectors, forming a two-dimensional feature matrix with a shape of $E = [N, E_s]$, where *Es* represents the predefined embedding size. This mapping effectively addresses the heterogeneity of sensor data, ensuring consistency in feature representation across different sensor sources.

Bilinear Attention: Bilinear attention is defined as $A = \text{Tan}h(EE^T)$, where *Tanh* represents the hyperbolic tangent function. This computation produces a matrix *A*, where each element quantifies the similarity relationship between two features. This design effectively extracts key features and enhances the model's understanding of feature interactions, thereby improving anomaly detection accuracy.

Permuted CNN: For the matrix A generated in the previous step, a permutation matrix *M* is designed to rearrange its elements, where originally adjacent elements are dispersed, while distant elements are brought closer together, resulting in a new matrix *P*, defined as $P \triangleq MAM^T$. Subsequently, CNNs are applied separately to both the original matrix *A* and the permuted matrix *P* to capture local and non-local relationships between matrix elements. Finally, feature concatenation and a fully connected layer are employed to achieve relation fusion. This design effectively extracts local feature correlations as well as long-range dependencies between features, thereby enhancing the model's robustness in anomaly detection.

Residual Connection: In the residual connection, a Multilayer Perceptron (MLP) is used as a supplementary pathway. The embedding matrix *E* is flattened and then mapped to the target space. The MLP extracts global feature information through fully connected layers, addressing the limitations of CNNs in global feature extraction. Subsequently, the output from the MLP is weighted and fused with the results from the Permute Convolution pathway to enhance the feature representation capability. This design effectively improves the stability of feature representation, thereby enhancing the robustness of the model in anomaly detection tasks.

Time-Series Processing Module: This module primarily focuses on capturing feature correlations in the temporal dimension of multi-source sensor data. First, Long Short-Term Memory (LSTM) networks are utilised to extract time-series patterns from individual sensor data, effectively capturing their dynamic variations. Next, the temporal correlations across all sensors are stacked, and the feature extraction process is further optimised through Bilinear Attention, Permute CNN, and Residual Connection. This approach fully explores the interactions among multi-source data over time, enhancing the model's understanding and modelling of temporal features.

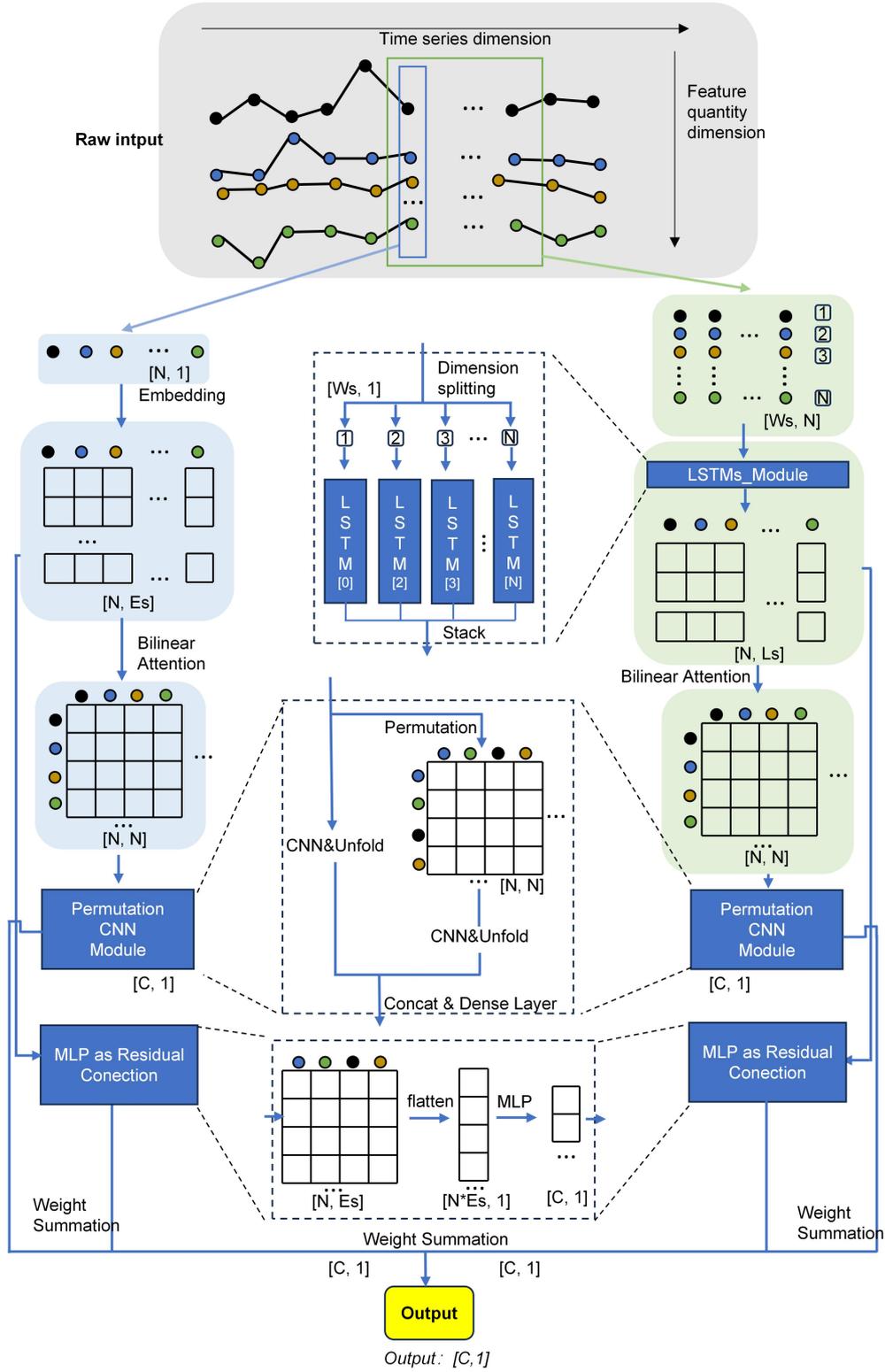

**Figure 2.** Schematic Diagram of the Time-EAPCR Model Results

LSTM Module：After the raw data is input, a historical time window for each feature is obtained based on a predefined window size, denoted as [$W_S$, N], where *Ws* represents the window length and *N* denotes the number of different sensor data streams.An LSTM-based feature extractor is then introduced, where each LSTM unit independently processes the time-series data from a single sensor. The *w*-dimensional windowed data is mapped to an *Ls*-dimensional high-dimensional space, with each processed feature represented as $[1, L_s]$, where Ls denotes the predefined LSTM hidden layer size.Finally, the processed data from all sensors are stacked to form a time-series feature matrix *Et* with a shape of $[N, L_s]$.

After the LSTM module extracts the time representation matrix *Et* for all individual features, the same enhanced bilinear attention mechanism and permute convolution operations as in the Multi-Sensor Feature Module are applied. This process captures the dynamic correlations in the temporal evolution of the sensor network. Consequently, the extracted time-domain features undergo multiple refinement iterations.Subsequently, the output vectors obtained from this module are weighted and fused with the output vectors from the Multi-Sensor Feature Module to generate the final prediction vector. The overall model architecture is illustrated in Figure 2.

## 2.3 Evaluation Metrics

Given the low proportion of anomalous data, relying solely on precision as an evaluation metric is inadequate. Precision tends to be biased in highly imbalanced datasets, as it is predominantly influenced by the majority class, failing to comprehensively reflect the model's ability to detect the minority class (anomalous data).To address this issue, we introduce the Area Under the Receiver Operating Characteristic Curve (AUC-ROC) as the primary evaluation metric. AUC evaluates the classification performance of the model by considering both the True Positive Rate (TPR) and the False Positive Rate (FPR), and it is computed using the following formula:

$$TPR = \frac{TP}{TP + FN} \tag{1}$$

$$FPR = \frac{FP}{FP + TN} \tag{2}$$

$$AUC = \int_0^1 TPR \, d(FPR) \tag{3}$$

Definition of AUC and Its Importance in Imbalanced Data, True Positive (TP) represents the number of correctly identified anomalous samples.False Negative (FN) refers to the number of missed anomalous samples.False Positive (FP) denotes the number of misclassified normal samples.True Negative (TN) corresponds to the number of correctly identified normal samples.The AUC value ranges from 0 to 1, where a value closer to 1 indicates superior model performance in distinguishing between positive and negative samples (anomalous and normal data).By introducing

AUC, we can more comprehensively assess model performance under class imbalance scenarios. In anomaly detection tasks, AUC provides a more reliable measure of the model's ability to identify minority-class instances, ensuring a robust evaluation of its detection capability.

$$F1_{Score} = 2 * \frac{Precison * Recall}{Precison + Recall} \tag{4}$$

This multi-metric evaluation strategy mitigates the limitations of using a single metric: AUC focuses on assessing the overall classification capability, while the F1-score emphasises the model's performance in recognising the minority class (anomalous samples) at a fixed threshold.Therefore, in the experimental section of this study, we primarily use AUC as the main evaluation metric, with F1-score as a supplementary measure.

## 3 Results

### 3.1 Main result

To systematically assess the generalisation capability of the Time-EAPCR model, this study conducted multi-dimensional experimental validation on four benchmark datasets covering natural ecosystems and urban infrastructure (Ponds, UWO S1/S2, and WaterHub).As shown in Table 2, compared with the baseline models evaluated by Russo et al. (Russo et al., 2021), Time-EAPCR demonstrates significant performance advantages across all test scenarios.

Specifically, on Dataset 1 (Ponds), Time-EAPCR achieved an AUC score of 0.9999, demonstrating exceptional performance. While other baseline models also attained an AUC of 0.9999, the F1-score at a fixed threshold further highlights the superiority of the proposed method. At the conventional threshold of 0.5, Time-EAPCR outperforms the competing models, reinforcing its effectiveness in anomaly detection.

On Dataset 2 (UWO S1), Time-EAPCR achieved an F1-score of 0.9912 and an AUC of 0.9567. Compared to the best-performing traditional ANN model, Time-EAPCR demonstrated a significant advantage in AUC performance, while also improving the F1-score by more than 0.14.

On Dataset 3 (UWO S2), the presence of a large number of complex collective anomalies posed significant challenges for traditional methods such as ANN, SVM, and RF, limiting their detection performance. Under baseline conditions, the best AUC achieved by conventional methods was only 0.8387.In contrast, Time-EAPCR demonstrated substantial performance improvements, achieving an AUC score of

0.9845 and an F1-score of 0.9, significantly outperforming the benchmark methods.

On Dataset 4 (WaterHub), Time-EAPCR further demonstrated its superiority, achieving an AUC score of 0.9831 and an F1-score of 0.856, both of which surpassed the performance of the baseline methods.Through experimental validation across multiple types of environmental datasets, the Time-EAPCR model exhibited outstanding anomaly detection performance in both natural ecological environments and urban environmental systems. These results confirm its generalisation capability and practical applicability across diverse and complex scenarios.

**Table 2.** Experimental Results on Public Datasets

| Models \ Datasets | Ponds | | UWO S1 | | UWO S2 | | Waterhub | |
|---|---|---|---|---|---|---|---|---|
| | AUC | F1 | AUC | F1 | AUC | F1 | AUC | F1 |
| DANB | 0.9773 | 0.5492 | 0.9328 | 0.6910 | 0.8185 | 0.5493 | 0.7466 | 0.372 |
| KNN | 0.9999 | 0.9548 | 0.9654 | 0.7864 | 0.8387 | 0.6 | 0.965 | 0.7591 |
| RF | 0.9997 | 0.9356 | 0.9677 | 0.829 | 0.8368 | 03606 | 0.9349 | 0.7282 |
| SVM | 0.9858 | 0.6995 | 0.9422 | 0.6873 | \ | \ | \ | \ |
| ANN | 0.9999 | 0.9443 | 0.9554 | 0.8023 | 0.8364 | 0.8364 | 0.9568 | 0.7526 |
| **Time-EAPCR** | **0.9999** | **0.9952** | **0.9913** | **0.9567** | **0.9845** | **0.9** | **0.9831** | **0.856** |

Figure 3 presents a comparative analysis of the ROC curves for the proposed method and baseline methods across multiple datasets. This experiment systematically evaluates model performance differences by aggregating false positive rates (FPR) and true positive rates (TPR) at varying classification thresholds for both approaches across all test datasets.In the figure, the dashed line represents the performance benchmark of a random classifier, while the solid curves illustrate the actual model performance. The distribution of the ROC curves shows that, in most cases, the ROC curve of the proposed method (red curve) is closer to the top-left quadrant compared to baseline methods. This indicates that the proposed approach achieves a higher true positive recognition rate while maintaining a lower false positive rate.

It is worth noting that on Dataset 1 (Ponds), both the baseline methods and the proposed method achieved AUC scores close to 1, with excellent ROC curve performance. This indicates that all methods demonstrated satisfactory global ranking capability, meaning the probability of assigning higher scores to positive samples than negative samples was consistently high across all approaches.

A high AUC score reflects the model's ability to distinguish between positive and negative samples in a relative sense, but it does not directly indicate classification performance at a specific threshold. In real-world applications, determining the optimal decision threshold is often challenging.As discussed earlier, this study evaluates the F1-score at a conventional threshold to provide a more practical assessment. Experimental results show that although all methods achieve similar AUC scores on Dataset 1, the proposed method outperforms baseline methods in terms of F1-score. This finding demonstrates that the proposed method not only exhibits superior global ranking capability but also possesses greater practical utility and robustness in real-world classification tasks.

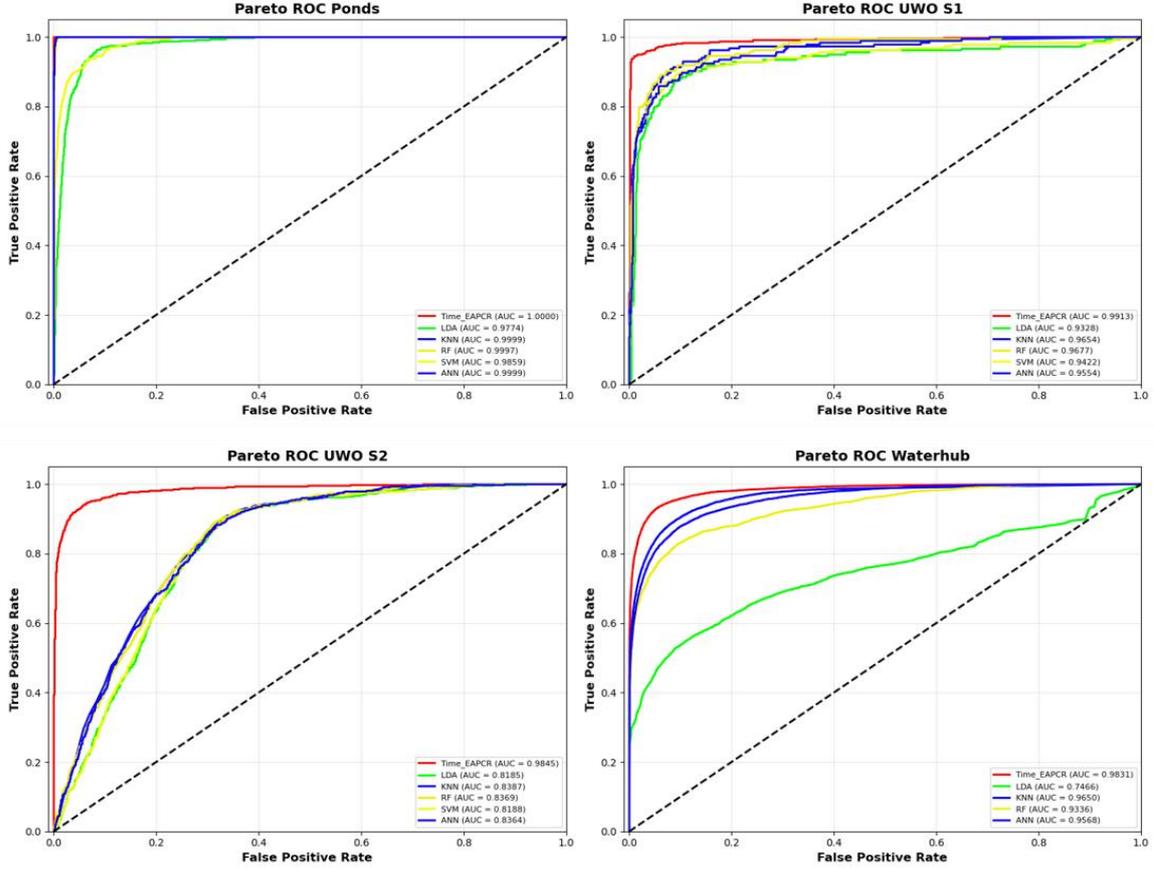

**Figure 3.** ROC Curves for Each Model on Public Datasets

## 3.2 Further analysis

In this section, we primarily explore the impact of the sliding window size for historical time observations on model performance, along with several key ablation experiments. These analyses aim to demonstrate the influence of model parameters on its final effectiveness and to validate the overall structural efficacy of the model.

### 3.2.1 Windowsize selection

In this experiment, we systematically investigated the impact of time window parameters on model performance by conducting a sensitivity analysis with five typical window sizes: 72, 96, 120, 168, 192. As shown in Table 3, the temporal dynamics of different environmental systems result in varying window optimisation strategies, highlighting the system-specific nature of the window size selection.

Dataset 1 (Ponds) demonstrates strong window robustness, with AUC scores remaining stable at 0.9999 across all window configurations. The 168-window configuration was ultimately selected as the baseline configuration, as it achieved the best performance with an F1-score of 0.9439, outperforming other window sizes.

Dataset 2 (UWO S1) achieves its performance peak with the 120-window configuration, where the AUC reaches 0.9913, and the F1-score attains a commendable value of 0.9567.

Dataset 3 (UWO S2) exhibits more complex collective anomaly features compared to UWO S1, necessitating the use of a broader range of time windows to capture more reliable time-series information. Experimental results show that the model performs better with the 168-window configuration.

Dataset 4 (WaterHub) achieves an AUC of approximately 0.98 across multiple window configurations, with the 120-window configuration providing the best performance, yielding both an excellent AUC and F1-score.

The experimental results indicate that the choice of time window has a significant impact on model performance, though the extent of this influence varies depending on the temporal resolution, anomaly types, and complexity of the dataset. Overall, despite some parameter sensitivity, the proposed model outperforms the baseline methods across all window configurations, demonstrating considerable anomaly detection capability. This provides crucial technical support for the engineering deployment of real-world environmental monitoring systems.

**Table 3.** Sensitivity Experiment Results of the Model for Different Historical Window Sizes Across Datasets

| Datasets \ Window sizes | 72 | | 96 | | 120 | | 144 | | 168 | | 192 | |
|---|---|---|---|---|---|---|---|---|---|---|---|---|
| | AUC | F1 | AUC | F1 | AUC | F1 | AUC | F1 | AUC | F1 | AUC | F1 |
| Ponds | 0.9999 | 0.9951 | 0.9999 | 0.993 | 0.9999 | 0.9941 | 0.9999 | 0.981 | 0.9999 | 0.9952 | 0.9999 | 0.9946 |
| UWO S1 | 0.9736 | 0.8888 | 0.9871 | 0.9326 | 0.9913 | 0.9567 | 0.9797 | 0.9186 | 0.9850 | 0.9245 | 0.9768 | 0.865 |
| UWO S2 | 0.955 | 0.7722 | 0.94 | 0.72 | 0.9379 | 0.7117 | 0.9753 | 0.8386 | 0.9845 | 0.9 | 0.9 | 0.635 |
| Waterhub | 0.9811 | 0.8 | 0.9787 | 0.8526 | 0.9831 | 0.856 | 0.9770 | 0.8425 | 0.9813 | 0.8633 | 0.9803 | 0.8547 |

### 3.3.2 Ablation study

In this section, we conducted ablation experiments. As mentioned in the methodology section, the proposed model consists of two primary components: the Multi-Sensor Feature Module and the Time-Series Processing Module. To validate the overall effectiveness of the model, we used the best-performing window configuration as the baseline. Ablation experiments were conducted to evaluate the performance of each module individually. As shown in Table 4, across all datasets, combining both modules consistently achieved the best AUC performance.

The Multi-Sensor Feature Module focuses solely on the correlation of data from different sensors at the same time point, while neglecting the dynamic changes in the temporal dimension. As a result, it struggles to accurately capture the evolution patterns of temporal features, leading to limited performance. In contrast, the Time-Series Processing Module prioritises temporal dimension information, excelling at extracting nonlinear dynamic interactions and temporal dependencies in environmental data.Across all four datasets, the Time-EAPCR model outperformed the results of using either module independently, as reflected in the AUC and F1 scores, thereby validating the effectiveness of combining both modules.

In Dataset 1 (Ponds), both modules individually achieved commendable performance, and their combination yielded an excellent overall result. In UWO S1 and S2, the results show that the Time-Series Processing Module contributes more significantly to the model's performance, with results clearly outperforming those from the Raw Module.Additionally, Dataset 3 (UWO S2) exhibits complex collective anomalies, and when the temporal dimension information is not considered, the model's performance is significantly poorer. However, when combined with the Time-Series Module, the model surpasses the performance of all single-module configurations, indicating the feasible complementary effect between the two independent modules.While this structure increases the overall complexity and parameter count of the model, we believe the improved results justify the trade-off.

**Table 4.** Ablation Experiment Results. Testing the Performance of the Two Key Modules Independently and the Overall Structural Effectiveness

| Modules / Datasets | Multi-Sensor Feature Module | | Time Series Module | | Time-EAPCR | |
|---|---|---|---|---|---|---|
| | AUC | F1 | AUC | F1 | AUC | F1 |
| Ponds | 0.9990 | 0.8982 | 0.9999 | 0.9925 | 0.9999 | 0.9952 |
| UWO S1 | 0.9296 | 0.7357 | 0.9879 | 0.9379 | 0.9913 | 0.9567 |
| UWO S2 | 0.8299 | 0.4776 | 0.9795 | 0.984 | 0.9845 | 0.9897 |
| Waterhub | 0.9397 | 0.74 | 0.9741 | 0.8242 | 0.9831 | 0.851 |

### 3.3 Real dataset result

To assess the potential applicability of the proposed method in real-world environments, we conducted experimental testing using real monitoring data from the Luoshi River. Specifically, historical data from October 31, 2020, to March 28, 2024, was used as the training dataset, while data from May 21, 2024, to October 31, 2024, served as the test dataset.As described in Section 2.1, based on the Chinese Surface Water Quality Standards and the annual water management reports for the Luoshi

River, water quality below Class V is considered as an anomalous pollution event for this river. This was used as the basis for experimenting with the proposed method.

The experimental results show that with a time window size of 6, the proposed method achieves an AUC of 0.9959, and the F1-score at the fixed threshold reaches 0.9249. Considering the coarse temporal resolution of the monitoring system, which collects data every 4 hours, a larger time window could introduce unnecessary noise. As shown in Table 5, setting the time window to 6 effectively meets the accuracy requirements for anomaly detection.

**Table 5.** Experimental Results on the Private Dataset Obtained from Real-World Monitoring Activities

| Window Sizes | 3 | | 6 | | 12 | | 24 | |
|---|---|---|---|---|---|---|---|---|
| | AUC | F1 | AUC | F1 | AUC | F1 | AUC | F1 |
| | 0.9861 | 0.7733 | 0.9959 | 0.9249 | 0.980 | 0.7001 | 0.9874 | 0.8246 |

As shown in Figure 4, the time-series monitoring data from the test set (May 20, 2024, to October 31, 2024), alongside the model's predictions, validates the practical applicability of Time-EAPCR. The study area, Dali City, located in the subtropical highland monsoon climate zone, experiences significant fluctuations in water quality parameters during the rainy season (June-November). The monitoring data reveals that anomalous water quality events primarily occur towards the end of the flood season in October, exhibiting a sustained anomaly pattern lasting from 7 to 21 days, contrasting with the discrete single-point anomalies observed earlier (May-September).The Time-EAPCR model, trained on two years of historical data, demonstrated strong detection capability in this complex climatic context:1) It effectively identified sustained compound anomalies and discrete single-point anomalies.2) The model maintained detection stability during the parameter fluctuation phase (F1-score > 0.92).This adaptability to non-stationary environmental systems confirms the practical value of the proposed method for water quality monitoring in climate-sensitive regions.

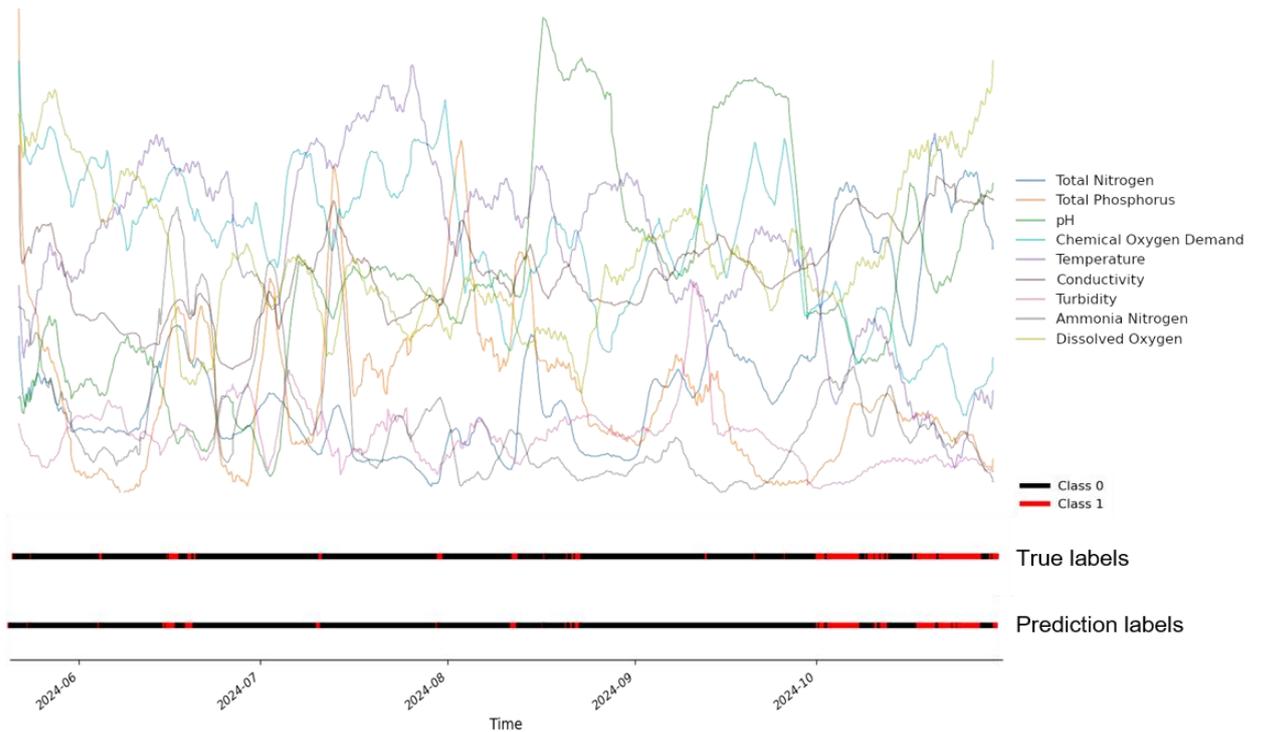

**Figure 4.** Visualisation of Original Data and Detection Results

# 4 Conclusion

This study proposes the Time-EAPCR framework, based on spatiotemporal feature fusion, which effectively addresses the challenges of anomaly detection in multi-source heterogeneous data within environmental monitoring. By establishing a collaborative mechanism between the in-situ feature interaction module and the time-series interaction module, the framework successfully performs accurate cross-sensor correlation feature detection for environmental system anomalies.

Experimental results demonstrate that the proposed method achieves satisfactory performance across typical scenarios, such as natural water bodies and urban infrastructure. Compared to traditional methods, the Time-EAPCR model not only improves detection accuracy but also exhibits strong generalisation capability, supporting its application in diverse environmental settings. Furthermore, experiments conducted in a real river system scenario show that the method effectively identifies both discrete anomalies and sustained anomalies over long-term monitoring activities, providing valuable insights for intelligent environmental system detection.

Overall, Time-EAPCR offers a powerful anomaly detection tool, presenting a more intelligent approach for environmental system monitoring. It effectively fills the gap in applying deep learning techniques to the intelligent diagnosis of complex environmental systems. Its high accuracy and strong transferability make it feasible for real-world deployment. In future work, we aim to further optimise the method's performance, with the potential to contribute to a breakthrough in the fully intelligent management of environmental systems within the rapidly advancing field of artificial intelligence.

## Data and Code Availability

The public datasets can be found in the corresponding references. The source code and private dataset can be made available upon reasonable request to the corresponding author.

## Acknowledgments

This work was supported by the National Natural Science Foundation of China (42367066, 62106033), Yunnan Fundamental Research Projects (202401AT070016, 202301BA070001-037), and National Observation and Research Station of Erhai Lake Ecosystem in Yunnan (2022ZZ01)

## Author contribution statement

Conceptualization, Chichun Zhou and Zhenyu Zhang; Experimental work, Lei Liu, Ling An, and Yuchao Lu; Formal analysis, Lei Liu, Ling An, and Huajie Liang; Investigation, Lei Liu, Ling An, and Huajie Liang; Writing - original draft, Lei Liu and Yuchao Lu; Funding acquisition, Chichun Zhou, Zhenyu Zhang; Writing-review & editing, Chichun Zhou, Zhenyu Zhang and Ling An; All the authors have read and agreed to the published version of the manuscript.

## Additional information

Conflicts of Interest: The authors declare that they have no known competing financial interests or personal relationships that could have appeared to influence the work reported in this paper.